%% file: main.tex
\documentclass[letterpaper, 10 pt, conference]{ieeeconf}  

\IEEEoverridecommandlockouts                              

\usepackage{graphics} 
\usepackage{graphicx}
\usepackage{epsfig} 
\usepackage{times} 
\usepackage{amsmath} 
\usepackage{amssymb}  
\usepackage[english]{babel}
\usepackage{amsthm}

\usepackage{bm}
\usepackage[linesnumbered,ruled,vlined]{algorithm2e}
\usepackage[noend]{algpseudocode}
\usepackage[font={small}]{caption}
\usepackage{subcaption}
\usepackage{physics}
\usepackage[table]{xcolor}
\usepackage{tikz}

\usetikzlibrary {arrows.meta}
\usetikzlibrary{fit,
                positioning}
\usepackage[belowskip=1ex]{subcaption}  
\usepackage{caption}
\usepackage[ruled,vlined]{algorithm2e}
\usetikzlibrary{calc} 
\usetikzlibrary{positioning}
\usepackage{balance}
\usepackage{soul}
\usetikzlibrary{arrows}
\usepackage{cite}
\usepackage{url}
\usepackage{multirow}
\usepackage{booktabs} 
\usepackage[hidelinks]{hyperref}
\usepackage{adjustbox}
\input{tex/preamble.tex}

\usepackage{makecell} 
\usepackage{setspace}
\title{\LARGE \bf
Language-Enhanced Mobile Manipulation for Efficient Object Search in Indoor Environments
}

\author{
Liding Zhang$^{1\ast}$, 
Zeqi Li$^{1\ast}$, 
Kuanqi Cai$^{1\ast}$, 
Qian Huang$^{1}$, 
Zhenshan Bing$^{2,1}$, 
Alois Knoll$^{1}$
\thanks{$^{1}$L. Zhang, Z. Li, K. Cai, Q. Huang, Z. Bing and A. Knoll are with the School of Computation, Information and Technology (CIT), Technical University of Munich, 80333 Munich, Germany.
{\tt\small liding.zhang@tum.de}}
\thanks{$^{2}$Z. Bing is also with the State Key Laboratory for Novel Software Technology and the School of Science and Technology, Nanjing University (Suzhou Campus), China.~\textit{(Corresponding author: Zhenshan Bing.)}}
\thanks{$^{\ast}$These authors contributed equally to this work.}
\thanks{The authors acknowledge the financial support by the Bavarian State Ministry for Economic Affairs, Regional Development and Energy (StMWi) for the Lighthouse Initiative KI.FABRIK (Phase 1: Infrastructure and the research and development program under grant no. DIK0249).}
}

\begin{document}

\maketitle
\thispagestyle{empty}
\pagestyle{empty}

\begin{abstract}

Enabling robots to efficiently search for and identify objects in complex, unstructured environments is critical for diverse applications ranging from household assistance to industrial automation. However, traditional scene representations typically capture only static semantics and lack interpretable contextual reasoning, limiting their ability to guide object search in completely unfamiliar settings. To address this challenge, we propose a language-enhanced hierarchical navigation framework that tightly integrates semantic perception and spatial reasoning. Our method, Goal-Oriented Dynamically Heuristic-Guided Hierarchical Search (GODHS), leverages large language models (LLMs) to infer scene semantics and guide the search process through a multi-level decision hierarchy. Reliability in reasoning is achieved through the use of structured prompts and logical constraints applied at each stage of the hierarchy. For the specific challenges of mobile manipulation, we introduce a heuristic-based motion planner that combines polar angle sorting with distance prioritization to efficiently generate exploration paths. Comprehensive evaluations in Isaac Sim demonstrate the feasibility of our framework, showing that GODHS can locate target objects with higher search efficiency compared to conventional, non-semantic search strategies. Website and Video are available at: \href{https://drapandiger.github.io/GODHS}{\textcolor{blue}{https://drapandiger.github.io/GODHS}}.

\end{abstract}

\input{sections/sec1}
\input{sections/sec2}
\input{sections/sec3}

\input{sections/sec4}
\input{sections/sec5}





\bibliographystyle{IEEEtran}
\bibliography{references}

\end{document}

%% file: tex/preamble.tex
\usepackage{amssymb}
\usepackage{amsfonts}
\usepackage{amsmath}
\usepackage{amsthm}
\usepackage{bm}

\input{tex/comments.tex}

\input{tex/symbols}

%% file: tex/comments.tex
\usepackage[normalem]{ulem}                                        
\usepackage{marginnote}
\usepackage[textwidth=10ex,colorinlistoftodos]{todonotes}

\colorlet{mh}{red}
\colorlet{fwu}{red}
\colorlet{ywu}{blue}
\colorlet{kchen}{blue}
\colorlet{lchen}{green}
\colorlet{zbing}{green}
\colorlet{shaddadin}{purple}
\colorlet{iperez}{cyan}
\colorlet{schneider}{magenta}



%% file: tex/symbols.tex
\usepackage{amssymb}
\usepackage{mathtools}
\usepackage{sansmath}

\mathchardef\mhyphen="2D   

\newcommand{\RNum}[1]{\uppercase\expandafter{\romannumeral #1\relax}}

%% file: sections/sec1.tex
\section{Introduction}

When humans search for objects in unfamiliar environments, they rely on a hierarchical understanding of semantic information to rapidly localize potential object placements~\cite{kuipers2000spatial}. Inspired by this ability, our approach emulates the human strategy of leveraging semantic cues—e.g., “pillows often lie on beds”—to guide the search process. Rather than exhaustively scanning an entire space, humans focus on “carriers” (like beds or tables) that are most likely to hold a target object, significantly reducing the search effort. 
Mobile manipulation has made great strides in recent years~\cite{ZHANG2025100207}. However, most robots still rely on purely spatial exploration strategies and ignore semantic cues, which makes them inefficient in unfamiliar environments.

\begin{figure}[!t]
    \centering
    \includegraphics[width=0.9\columnwidth]{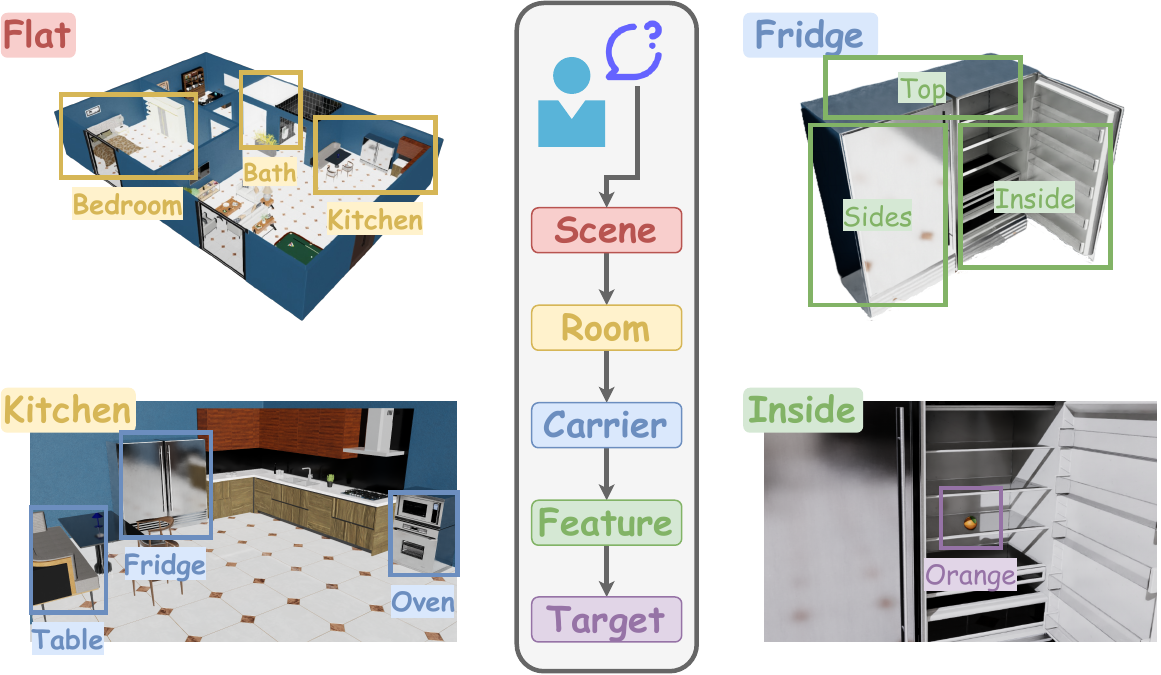}
    \caption{The \textbf{GODHS framework} divides each scene into five strict levels: Scene, Room, Carrier, Feature, and Item. For instance, after entering the \textbf{\textit{flat}}, the robot uses mapping and semantic segmentation to determine and prioritize the types of rooms through a LLM. For the first room \textbf{\textit{kitchen}}, the LLM classifies and prioritizes carriers such as furniture or devices. For the first carrier \textbf{\textit{fridge}}, the LLM identifies and prioritizes the feature regions worth searching. Finally, for the first feature \textbf{\textit{inside}}, the robot searches for the presence of the target object \textbf{\textit{orange}}.}
    \label{fig:1}
    \vspace{-1.5em}
\end{figure}

In addition, current navigation systems exhibit two major shortcomings. First, although they may incorporate basic semantic labels, they often fail to reason about the relationships between known objects and unknown targets. Second, while vision-language models (VLMs) can align language with image features, they often lack the broad commonsense knowledge that large language models (LLMs) possess. In principle, LLMs combined with robust semantic segmentation could provide deeper real-world contextual reasoning.

To address these gaps, this paper introduces the \emph{Goal-Oriented Dynamically Heuristic-Guided Hierarchical Search} (\textbf{GODHS}) framework, an exploratory study inspired by human search behavior. Building on the notion that people mentally decompose a search task—from room, to carrier, to specific features—GODHS employs LLMs to orchestrate a five-level search hierarchy: \emph{scene} $\rightarrow$ \emph{room} $\rightarrow$ \emph{carrier} $\rightarrow$ \emph{feature} $\rightarrow$ \emph{item}. As illustrated in Fig.~\ref{fig:1}, each level narrows the search scope through an LLM-driven reasoning process.

Implementing this hierarchical pipeline presents two primary challenges that our work addresses:
\begin{itemize}
    \item \textbf{Reliable Hierarchical Reasoning:} The system must translate a high-level search goal into a multi-step, semantically-grounded plan. GODHS achieves this by leveraging LLMs to hierarchically decompose the problem. To ensure the reliability of the LLM's guidance, we employ structured prompts and logical constraints at each level of the hierarchy.
    
    \item \textbf{Efficient Physical Exploration:} A mobile manipulator exploring a carrier faces a vast action space for positioning its base and end-effector. We propose a heuristic-based motion planner that combines polar and lexicographical sorting to efficiently generate and prioritize exploration poses, significantly reducing redundant travel and execution time.
\end{itemize}

%% file: sections/sec2.tex
\section{Related Work}
\label{sec:related_work}

Robotic object search has progressed through three main paradigms. Early geometric approaches established principles but faced critical limitations. These methods constrained search regions using spatial affordances~\cite{10.1007/BF01421203} or modeled uncertainty with probabilistic frameworks~\cite{zhang2025TASE, GELENBE1998319,zhang2025apt}. However, they were often effective only in structured environments, as they required exhaustive object relationship priors and could not infer latent functional associations (e.g., that medicine belongs in a cabinet). Underpinning these systems are foundational motion planning algorithms~\cite{zhang25ral}, which have evolved from sampling-based approaches to adapted high-DoF robotic systems for real-time planning~\cite{cai2025demostration, ZHANG2025git}. Furthermore, traditional decision-making frameworks like POMDPs~\cite{7759839} remained confined to predefined taxonomies, limiting their adaptability to novel objects.

The advent of VLMs enabled open-vocabulary object recognition but introduced new constraints. While semantic mapping systems~\cite{patki2019languageguidedsemanticmappingmobile} enriched spatial maps with object labels, they often treated semantics as static attributes, overlooking dynamic relational reasoning. Other works used hierarchical graphs to model object interactions~\cite{Remote_object}, but these rigid structures typically required complete environmental pre-knowledge. Modern VLM-integrated navigation methods~\cite{huang2023visuallanguagemapsrobot, chang2023goatthing} can locate open-vocabulary targets but may overfit to superficial visual-text correlations~\cite{tang2025openinopenvocabularyinstanceorientednavigation}, struggling to chain contextual relationships for multi-step reasoning.

Recent integrations of LLMs show strong potential but still face challenges. LLM-augmented planners generate plausible search hypotheses~\cite{song2023llmplannerfewshotgroundedplanning}, yet many rely on flat decision structures that inefficiently evaluate object relationships. Scene-graph methods capture relationships more dynamically~\cite{Honerkamp_2024,Zhang2024Elliptical} but often lack incremental refinement and assume high observability~\cite{POMP}. While progress has been made in language grounding, few methods emulate human-like hierarchical task decomposition. Our work bridges this divide by synergizing LLM-driven commonsense reasoning with principled hierarchical action planning, avoiding both the inflexibility of purely geometric heuristics and the inefficiency of flat neural architectures.

%% file: sections/sec3.tex
\section{Methodology}

\begin{figure*}[htbp]
    \centering
    \includegraphics[width=1\textwidth]{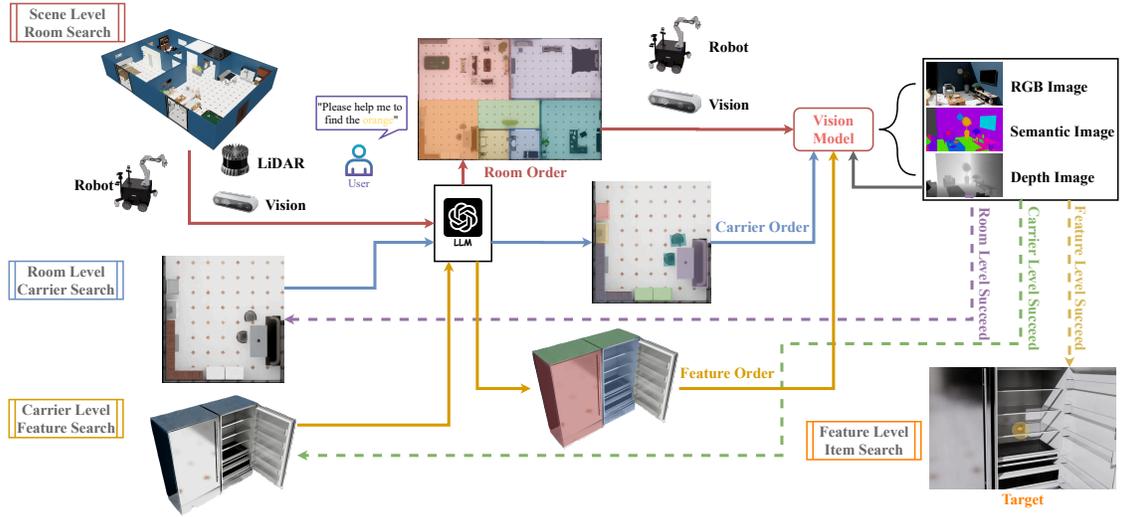} 
    \caption{\textbf{Complete Architecture of the GODHS System:} First, the user provides the target object through natural language input, which is then processed by a LLM to extract semantic intent. The robot captures geometric data using LiDAR and RGB images via cameras, generating a topological map with room names through the language model. Within the known map, the LLM-driven prioritization ranks rooms based on the likelihood of containing the target object and navigates to them sequentially. Within each room, the robot detects candidate objects through visual object detection, then utilizes the LLM to analyze whether they are carriers and ranks all collected carriers by the probability of containing the target. For each carrier, the LLM plans a hierarchical search strategy: first analyzing geometric features to identify subregions worth searching (top, interior, etc.), then prioritizing and searching these subregions according to probabilities assigned by the LLM. The loop terminates when the LLM verifies target recognition through visual-language grounding of sensor data.}
    \label{fig2}
    \vspace{-1.7em}
\end{figure*}

Our search approach follows a logical hierarchical progression to locate the ultimate target. We construct a hierarchically expandable algorithm, GODHS, that integrates environmental perception with commonsense reasoning from a large language model (LLM) to derive semantic action sequences(Sec.~\ref{sec3.1}). To ensure the reliability of the LLM-driven reasoning, we employ structured prompts and constraints at each stage of the hierarchy. To ensure the mobile robot advances sequentially around the carrier, we employ dictionary mapping and polar angle sorting (Sec.~\ref{sec3.3}).

\subsection{GODHS Framework}\label{sec3.1}

Goal-Oriented Dynamically Heuristic-Guided Hierarchical Search is an efficient search approach that combines cognitive reasoning, sensory data processing, and decision-making strategies. The detailed process is shown in Figure~\ref{fig2}. The GODHS algorithm is designed primarily to locate target objects in complex environments. It has four key features:

\begin{itemize}
    \item \textbf{Goal-Oriented:} The search is consistently focused on the final target, allowing the system to effectively prune the search space at each level of the hierarchy.
    
    \item \textbf{Dynamically Updated:} The system dynamically reassesses and reorders search priorities based on new information gathered during execution, inspired by dynamic tree search structures~\cite{SLEATOR1983362}.
    \item \textbf{Heuristic-Guided:} The process is guided by contextual probabilities from the LLM, which infers likely search locations based on commonsense knowledge rather than strict mathematical optimization.
    
    \item \textbf{Bounded Hierarchical:} The search space is organized into a five-level hierarchy (Scene $\rightarrow$ Room $\rightarrow$ Carrier $\rightarrow$ Feature $\rightarrow$ Item), progressively narrowing the scope to avoid an exhaustive global search~\cite{6507635}.
\end{itemize}

\begin{algorithm}[t!]
\caption{ObjectSearchGODHS($\mathsf{s}$, $\mathsf{t}$)}
\label{alg:1}
\DontPrintSemicolon
\small
\SetKwInOut{Input}{Input}
\SetKwInOut{Output}{Output}

\Input{$\mathsf{s}$ --- scene name, $\mathsf{t}$ --- target name}
\Output{$\mathsf{\tau}$ --- found target}
$\mathsf{\tau} \gets \textbf{False}$, $\mathcal{M_S},\mathcal{M_R},\mathcal{M_C} \gets \varnothing$, $\mathbf{R},\mathbf{C},\mathbf{F} \gets [\,]$\;
EnterScene($\mathsf{s}$)\;
\While{not IsSceneMapComplete($\mathcal{M_S}$)}{
  EnterRandomRoom()\;
  $\mathcal{M_R} \gets$ LidarToMap(LidarData())\;
  $\mathcal{M_S} \gets$ UpdateSceneMap($\mathcal{M_S}$, $\mathcal{M_R}$)\;
  $\mathcal{R},\mathcal{I_R} \gets$ RoomMap($\mathcal{M_R}$, InferRoom(SemSeg(CameraData())))\;
}
$\mathbf{R} \gets$ SortRooms($\mathcal{R}, \mathsf{t}$)\;
\ForEach{$\mathsf{r} \in \mathbf{R}$}{
  MoveToRoom($\mathsf{r}$, $\mathcal{M_R}$, $\mathcal{I_R}$)\;
  $\mathcal{C} \gets$ ClassifyCarrier(SemSeg(CameraObservation()))\;
  $\mathcal{M_C},\mathcal{I}_C \gets$ GetCarrierPCL($\mathcal{C}$, CarrierObservation())\;
  $\mathbf{C} \gets$ SortCarriers($\mathcal{C}, \mathsf{t}$)\;
  \ForEach{$\mathsf{c} \in \mathbf{C}$}{
    $\mathcal{F} \gets$ \{`top', `bottom', `sides', `inside'\}\;
    $\mathbf{F} \gets$ ReasonFeatures($\mathsf{t}$, $\mathcal{F}$)\;
    \ForEach{$\mathsf{f} \in \mathbf{F}$}{
        $\mathcal{M_F} \gets$ PredictFeatureMap($\mathcal{M_C}$, $\mathcal{I}_C$, $\mathsf{f}$)\;
        $\mathcal{P_{EE}} \gets$ DetermineEEPoses($\mathcal{M_F}$)\;
        $\mathcal{P_{CH}} \gets$ DetermineCHPoses($\mathcal{P_{EE}}$, $\mathcal{M_F}$, $\mathcal{I_F}$)\;
        $\mathcal{P_{CH}^{EE}} \gets$ CHToEEPoses($\mathcal{P_{EE}}$, $\mathcal{P_{CH}}$, $\mathcal{M_F}$, $\mathcal{I_F}$)\;
        $\mathcal{P_{\text{CH}}^{\text{EE}}} \gets$ PosesSorting($\mathcal{P_{CH}^{EE}}$)\;
      \ForEach{$\mathbf{P_{\text{CH}}} \in \mathcal{P_{\text{CH}}}$}{
        NavigateToCHPose($\mathbf{P_{\text{CH}}}$)\;
        \ForEach{$\mathbf{P_{\text{EE}}} \in \mathcal{P_{\text{EE}}}$}{
          NavigateToEEPose($\mathbf{P_{\text{EE}}}$)\;
          \If{$\mathsf{t} \in$ SemSeg(CameraData())}{
            \Return \textbf{True}\;
          }
        }
      }
    }
  }
}
\Return \textbf{False}\;
\end{algorithm} 

The full process is detailed in Algorithm~\ref{alg:1}, which takes a scene name $\mathsf{s}$ and a target $\mathsf{t}$ as input. The algorithm initializes the required maps ($\mathcal{M_S}, \mathcal{M_R}, \mathcal{M_C}$) and lists ($\mathbf{R}, \mathbf{C}, \mathbf{F}$). The process begins with an autonomous exploration phase (Lines 2-8), where the robot constructs a global scene map $\mathcal{M_S}$ by sequentially visiting and mapping all accessible rooms. For each room, a local map $\mathcal{M_R}$ is built using LiDAR data and integrated into the global map: $\mathcal{M}_S \leftarrow \mathcal{M}_S \cup \left( \mathcal{M}_R \setminus \mathcal{M}_S \right)$.

The function $\text{InferRoom}(\cdot)$ in Line 7 employs an LLM to predict the room category $\mathbf{r}^*$ based on the set of observed objects $\mathcal{O}$:
\begin{equation}
    \mathbf{r}^* = \underset{\mathsf{r} \in \mathcal{KB}}{\arg\max} \, P\left(\mathsf{r} \mid \mathcal{O}\right),
\end{equation}
where $\mathcal{KB}$ represents the commonsense knowledge of the pre-trained LLM. This generates a room-to-map correspondence $\mathcal{I_R}: \mathcal{R} \to \mathcal{M_R}$. Finally, the language model ranks elements of $\mathcal{R}$ by likelihood of target $\mathsf{t}$ in $\mathcal{R}$:  
\begin{equation}\label{eq2}\mathbf{R} = \underset{\mathsf{r} \in \mathcal{R}}{\text{argsort}} \ P(\mathsf{r}\mid\mathcal{R},\mathsf{t}),\end{equation}

The agent navigates to rooms according to the order in $\mathbf{R}$. Within each room, the $\text{ClassifyCarrier}(\cdot)$ function in Line 11 prompts the LLM to identify which of the observed objects are plausible `carriers' for the target $\mathsf{t}$. This step directly yields a filtered list of carrier objects $\mathcal{C} \subseteq \mathcal{O}$ without relying on an explicit numerical threshold. Focused scanning to capture carrier point clouds $\mathcal{M}_C$ is performed and carrier-to-map correspondence $\mathcal{I}_C: \mathcal{C} \rightarrow \mathcal{M}_C$ will be established. These identified carriers are then ranked by the LLM based on their relevance to the target $\mathsf{t}$:
\begin{equation}\mathbf{C} = \underset{\mathsf{c} \in \mathcal{C}}{\text{argsort}} \ P(\mathsf{c}\mid\mathcal{C},\mathsf{t}).\end{equation}

Guided by the prioritized carrier list $\mathbf{C}$, the agent sequentially inspects each carrier $\mathsf{c} \in \mathbf{C}$. At the Feature Level, the agent must determine which specific parts of the carrier are most relevant for finding the target $\mathsf{t}$. We prompt the LLM to perform a direct selection and ranking task. Given a set of spatial regions $\mathcal{F} = \{\text{`top'}, \text{`bottom'}, \text{`sides'}, \text{`inside'}\}$, the LLM is tasked to return an ordered list of the most plausible features to inspect for the given carrier and target. This process yields a final, prioritized sequence of features $\mathbf{F} = \underset{\mathsf{f} \in \mathcal{F}}{\text{argsort}} \, P(\mathsf{f} \mid \mathcal{F}, \mathsf{t})$. The final stage of the algorithm~\ref{alg:1} in Lines 17-28 then iterates through this robustly generated list $\mathbf{F}$, creating and executing motion plans to visually inspect each feature in order, until the target is found.

\begin{figure}[htbp]
    \vspace{-0.5em}
    \centering
    \includegraphics[width=1\linewidth]{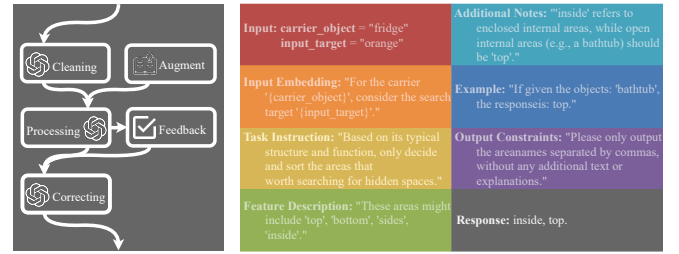}
    \caption{\textbf{LLM Reasoning Process (Left):} A multi-stage process involving data cleaning, processing with optional knowledge augmentation, and correcting with optional feedback ensures reliable output. \textbf{Prompt Design (Right):} An example of a structured prompt used to ground the LLM's reasoning for a specific task.}
    \label{fig3}
    \vspace{-1em}
\end{figure}

To ensure the reliability of the LLM's guidance within the GODHS framework, we must address the model's inherent tendency for statistical hallucination~\cite{ouyang2022traininglanguagemodelsfollow}. Inspired by recent structured reasoning techniques like Chain-of-Thought~\cite{wei2023chainofthoughtpromptingelicitsreasoning} and self-refinement~\cite{madaan2023selfrefineiterativerefinementselffeedback}, we implement a multi-stage verification process for LLM queries. This process, illustrated in Fig.~\ref{fig3} (Left), involves cleaning and structuring the input, executing the core reasoning task, and correcting the output to ensure it is semantically consistent and syntactically valid for the robot.

The cornerstone of this approach is a carefully structured prompt design, as shown in Fig.~\ref{fig3} (Right). For example, to determine the searchable features for a `fridge' potentially containing an `orange', the prompt explicitly defines the task, constrains the possible outputs to a predefined set (`top', `bottom', `sides', `inside'), provides clarifying examples (e.g., a `bathtub' should return `top'), and enforces a strict, machine-readable output format. This structured prompting is crucial for grounding the LLM's abstract knowledge to the specific, operational needs of the robot at each level of the hierarchy.

\subsection{Heuristic-Based Pose Generation and Sorting}\label{sec3.3}

\begin{figure*}[htbp]
    \centering
    \includegraphics[width=1\linewidth]{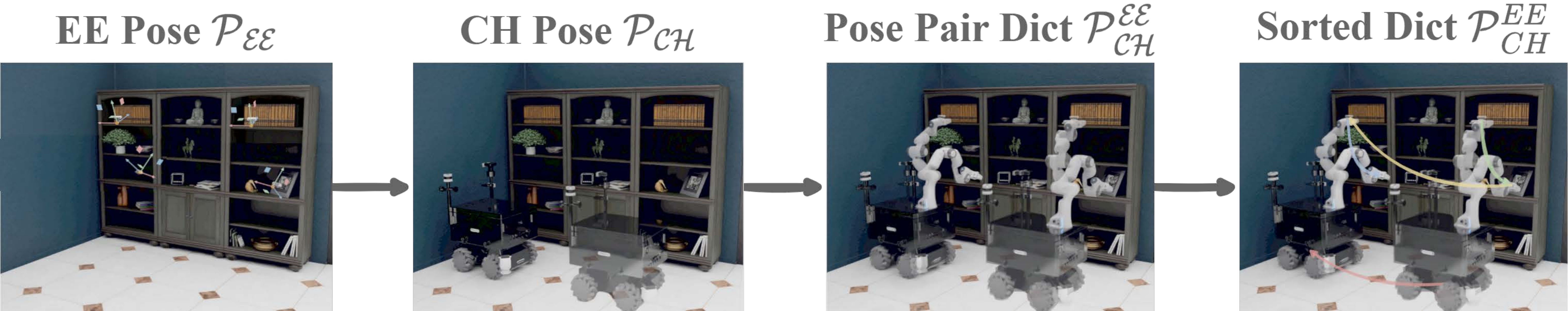} 
    \caption{\textbf{Leftmost: }Determine EE poses via carrier geometry analysis. \textbf{Second Left: }Generate CH poses through greedy EE pose exploration. \textbf{Second Right: }Verify CH-EE pairs with inverse kinematics validation. \textbf{Rightmost: }Prioritize CH poses by Polar Angle Sorting and EE poses by Lexicographical Sorting.}
    \label{fig4}
    \vspace{-1em}
\end{figure*}

This subsection details our heuristic-based methodology for computing chassis (CH) and end-effector (EE) poses. The goal is to efficiently generate structured search trajectories that enable the robot to systematically inspect a carrier's features while reducing redundant motion. The process involves five sequential stages: (1) extraction of geometric features \(\mathcal{M}_F\) from point cloud data, (2) generation and selection of EE poses \( \mathcal{P_{EE}} \), (3) fast evaluation and selection of chassis poses \( \mathcal{P_{CH}} \) via geometric solution, (4) validation of the relation between CH and EE \( \mathcal{P_{CH}^{EE}} \) with subsequent inverse kinematics (IK) solving, and (5) sorting of the resulting pose mappings \( \mathcal{P_{\text{CH}}^{\text{EE}}} \) for sequential execution.

The process begins with the acquisition of the point cloud of single carrier \(\mathcal{M}_C \subset \mathbb{R}^3\) and the occupied grid of the carrier \(\mathcal{M}_C^* \subset \mathbb{R}^2\). These data sources are used to extract the carrier's relevant features, collectively denoted as \(\mathcal{M}_F\). The features are subdivided into four components: Top Surface, Side Surface, Bottom Area and Inside Area.

The top surface feature \(\mathcal{M}_F^{\text{top}}\) is extracted directly from point cloud \(\mathcal{M}_C\). For each \((x_F,y_F)\) coordinate in the grid, the top surface point is identified as the point with the maximum \(z_F\) value. Formally, if we denote by \(\mathcal{M}_C^*\) the grid cell corresponding to a particular \((x_F,y_F)\) location, then 
\begin{equation}\hspace{-0.2em}
    \mathcal{M}_F^{\text{top}} = \left\{(x_F,y_F,z_F)\,\Big|\, 
        \begin{array}{l}
        (x_F,y_F) \in \mathcal{M}_C^*, \\
        (x_F,y_F,z_F') \in \mathcal{M}_C, \\
        z_F = \\\max\{z_F' \mid (x_F,y_F,z_F')\}
        \end{array}
        \right\}.
\end{equation}

The side surface feature \(\mathcal{M}_F^{\text{sides}}\) is derived from the boundary of the volumetric occupied grid \(\mathcal{M}_C^*\). For each \((x_F,y_F)\) coordinate lying on the topologically defined edge of \(\mathcal{M}_C^*\) (denoted by \(\partial \mathcal{M}_C^*\)), the corresponding side surface value is determined by extracting the maximum \(z_F\) value in \(\mathcal{M}_C\) over the range starting from \(z_F=0\). This could be expressed as: 
\begin{equation}
    \mathcal{M}_F^{\text{sides}} = \left\{ (x_F,y_F,z_F) \,\Big|\, 
\begin{array}{l}
(x_F,y_F) \in \partial \mathcal{M}_C^*, \\
(x_F,y_F,z_F') \in \mathcal{M}_C, \\
z_F \in \\\hspace{-0.8em}\left[0, \max\{z_F' \mid (x_F,y_F,z_F')\}\right]
\end{array}
\right\}.
\end{equation}

This representation captures the vertical boundaries of the carrier. The bottom area feature \(\mathcal{M}_F^{\text{bottom}}\) is defined by the intersection of the occupied grid \(\mathcal{M}_C^*\) with a spatially constrained horizontal plane at a fixed height \(z_F = z_{F0}\): 
\begin{equation}
    \mathcal{M}_F^{\text{bottom}} = \left\{ (x_F, y_F, z_{F0}) \,\Big|\, (x_F,y_F) \in \mathcal{M}_C^* \right\}.
\end{equation}

The inside feature is identified using a dedicated geometric analysis procedure designed to detect potential openings or enclosed spaces on the carrier.

For the mobile manipulator, two types of poses are essential: the chassis pose and the EE pose. The candidate chassis pose is defined as $\mathbf{p}_{CH}^{Can} = (x_{CH}, y_{CH}, \theta_{CH})$, where \((x_{CH}, y_{CH})\) represents the base position and \(\theta_{CH}\) the yaw angle. The candidate EE pose is defined as $\mathbf{p}_{EE}^{Can} = (x_{EE},y_{EE},z_{EE},\phi_{EE},\theta_{EE},\psi_{EE})$. The direction vector of the camera's view can be expressed as $\mathbf{v}_{\mathrm{dir}}=(\cos(\theta_{EE})\cdot\cos(\psi_{EE}),\cos(\theta_{EE})\cdot\sin(\psi_{EE}),\sin(\theta_{EE}))$. The vector pointing from the camera to the surface point is $\mathbf{v}_{\mathrm{point}}=\begin{pmatrix}x_F-x_{EE},y_F-y_{EE},z_F-z_{EE}\end{pmatrix}$. The surface point is considered covered by the vision cone of the camera, if the conditions of horizontal and vertical angle $\theta_{\text{horizontal}} = \arccos \left( \frac{v_{\text{dir},x} \cdot v_{\text{point},x} + v_{\text{dir},z} \cdot v_{\text{point},z}}{\| v_{\text{dir}} \| \| v_{\text{point}} \|} \right) < \text{FoV}_{\text{horizontal}}/2$ and $\theta_{\text{vertical}} = \arccos \left( \frac{v_{\text{dir},y} \cdot v_{\text{point},y} + v_{\text{dir},z} \cdot v_{\text{point},z}}{\| v_{\text{dir}} \| \| v_{\text{point}} \|} \right) < \text{FoV}_{\text{vertical}}/2$ are satisfied,  and the candidate pose earns one point. We employ a greedy algorithm~\cite{dijkstra2022note} as a heuristic to select a set of EE poses $\mathcal{P_{EE}}$ that provide sufficient visual coverage.

The set of chassis poses \(\mathcal{P_{CH}}\) is selected from \(\mathbf{p}_{CH}^{Can}\) via deterministic greedy algorithm based on \(\mathcal{P_{EE}}\). If a geometric solution can establish a collision-free connection, one point is awarded. The selection continues until \(\mathcal{P_{CH}}\) is found that spatially covers all elements in \(\mathcal{P_{EE}}\). 

Since fast-solving cannot guarantee a feasible solution for the actual robot operation, and directly solving the IK for all poses would lead to excessive computational overhead, we perform IK solving with Levenberg–Marquardt algorithm \cite{nakamura1986inverse} for each chassis pose $\mathbf{P_{\text{CH}}}$ in \(\mathcal{P_{CH}}\) and each end-effector pose $\mathbf{P_{\text{EE}}}$ in \(\mathcal{P_{EE}}\). If numerically stable solutions exist, the mapping is added to the dictionary \( \mathcal{P_{CH}^{EE}} \).  

The dictionary \( \mathcal{P_{CH}^{EE}} \) is non-sequential, so a sorting method is required to obtain an ordered version \(\mathcal{P}_{\text{CH}}^{\text{EE}}\). For each end-effector pose corresponding to a chassis pose, we apply Lexicographical Sorting based on the sequence \(z_{EE}, y_{EE}, x_{EE}, \psi_{EE}, \theta_{EE}, \phi_{EE}\).  

For all chassis poses, we do not simply sort based on spatial distance but instead ensure movement proceeds clockwise around the carrier. To achieve this, we use centroid-aligned Polar Angle Sorting \cite{Graham1972AnEA}, and the geometric centroid $(\bar{x},\bar{y})=\left(\frac{1}{n}\sum_{i=1}^{n}x_i,\frac{1}{n}\sum_{i=1}^{n}y_i\right)$ is computed where $(x_i,y_i)\in\mathcal{M}_{C}$, with $n$ being the size of the point set. Then we can calculate its angle relative to the average point by: 
\begin{equation}
\rho = \arctan_2\left( y_{\text{CH}} - \bar{y}, \, x_{\text{CH}} - \bar{x} \right),
\label{eq:polar_angle}
\end{equation} 
where $(x_{\text{CH}}, y_{\text{CH}})$ is the position of each chassis. Following the angle $\rho$, the chassis poses will be sorted. Using this method, the movement platform can advance around the carrier while steadily maintaining its direction of movement, avoiding the situation where sorting based on the nearest spatial distance might result in the actual path distance being significantly greater than the spatial distance.

%% file: sections/sec4.tex
\section{Experimental Setup and Results}

\begin{figure*}[htbp]
    \centering
    \includegraphics[width=0.98\textwidth]{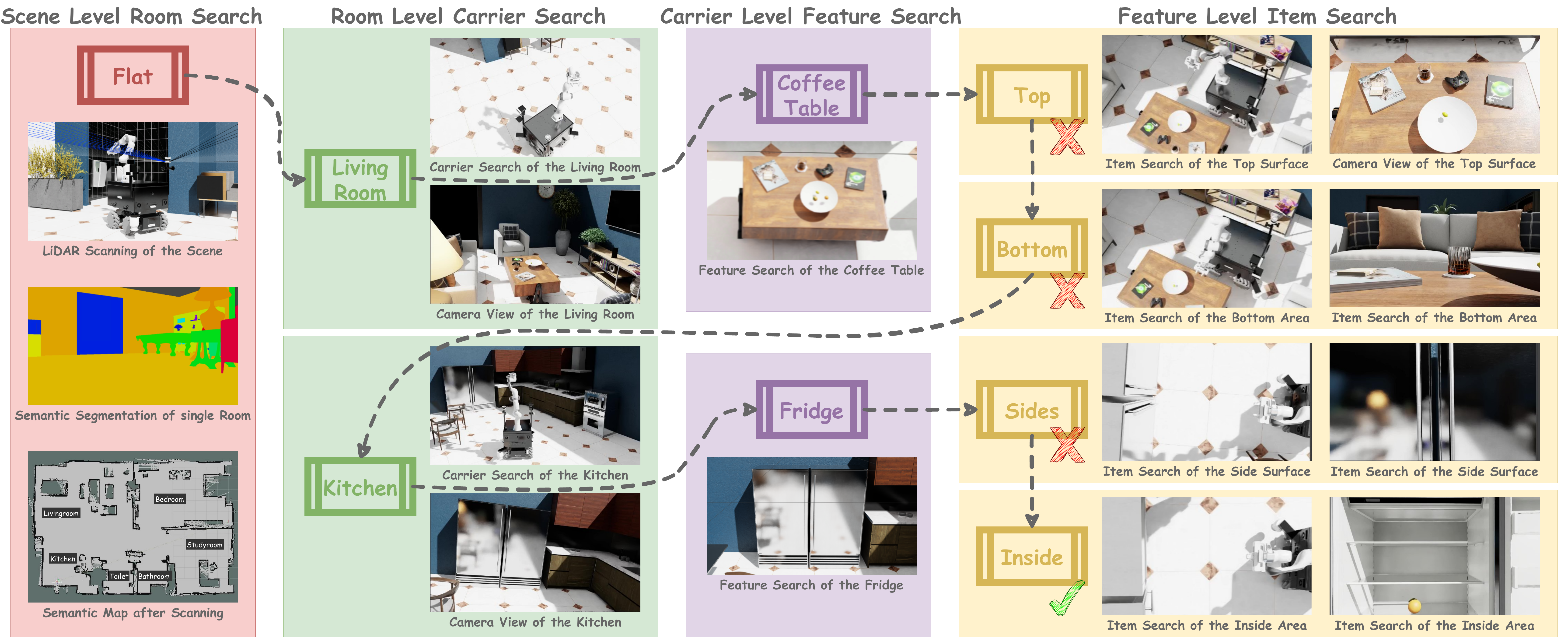}
    \caption{\textbf{Feasibility Example:} Taking a Qwen2.5-7B-powered search as an example, the target is an orange. After an initial exploration to map the scene (lower left), the LLM guides the robot to the living room first. It inspects the coffee table's top and bottom surfaces without success. It then proceeds to the kitchen, inspects the side of the fridge, and finally opens the door to find the orange inside, successfully completing the task.}
    \label{fig5}
    \vspace{-1em}
\end{figure*}

\subsection{Evaluation system setup}\label{sec4.1}


The simulation framework is developed using NVIDIA Isaac Sim, offering photorealistic sensor data and accurate physics simulation. The robotic platform is DARKO, featuring an omnidirectional RB-KAIROS base and a 7-DOF Franka Emika Panda arm, integrated with an Ouster OS1 LiDAR and an Intel RealSense D435 camera. The system runs on ROS Noetic, employing the standard Navigation Stack for base control and MoveIt for manipulator planning. A local Ollama server powers the cognitive layer for LLM-based interaction.

\subsection{Feasibility Testing}\label{sec4.2}


The feasibility test evaluates the system’s target search and semantic planning capabilities in a constrained indoor environment. Experiments were conducted in a simulated “flat” scene with seven functional zones (e.g., kitchen, living room, bedroom), as shown in Fig.~\ref{fig5}. A representative task involves locating an “orange” placed inside a fridge, requiring the robot to navigate across rooms and identify relevant carriers and features.
The process begins with environment exploration to build a complete occupancy map. Semantic segmentation detects objects, which are then processed by the LLM to infer room categories (e.g., recognizing a bed and dressing table suggests a bedroom). Based on the query “orange,” the LLM prioritizes searching the living room followed by the kitchen. In the living room, the robot inspects the coffee table’s top and bottom but finds nothing. It then moves to the kitchen, identifies the fridge as the most likely carrier, and successfully locates the orange inside.

To evaluate the search efficiency of our approach, we conducted 81 experiments using Qwen2.5-7B and GPT-4o. We define search efficiency metrics: the Room Search Rate ($R_r$), Carrier Search Rate ($R_c$), and Item Search Rate ($R_i$), which represent the percentage of rooms, carriers, or items, respectively, that were searched before the target was found. A lower value for these metrics indicates a more efficient search. The Overall Search Rate (OSR) is a weighted average of these values, representing the overall search cost:
\begin{equation}
    OSR = w_1 \cdot R_r + w_2 \cdot R_c + w_3 \cdot R_i,
\end{equation}
where \(w_1\), \(w_2\), and \(w_3\) are weights that reflect the relative importance of each search level. A reasonable choice for the weights could be assigned as $w_1 = 0.2$, $w_2 = 0.3$, $w_3 = 0.5$.

We compare our method against two traditional non-semantic search methods: a full Coverage Search and a Random Walk search. The results are shown in Table~\ref{tab1}.

\begin{table}[htbp]
    \centering
    \caption{Search efficiency of different strategies. Lower values indicate higher efficiency (less of the environment was searched).}
    \label{tab1}
    \begin{adjustbox}{center}
    \begin{tabular}{ccccc}
        \toprule
        \textbf{Method} & \textbf{$R_r$ (\%)} & \textbf{$R_c$ (\%)} & \textbf{$R_i$ (\%)} & \textbf{OSR (\%)} \\ \midrule
        \textbf{GPT-4o} & \textbf{21.43} & 20.53 & 21.17 & \textbf{21.03} \\ 
        \textbf{Qwen2.5} & 33.85 & \textbf{19.91} & \textbf{19.74} & 22.61 \\ 
        \textbf{Coverage} & 58.57 & 61.71 & 60.56 & 60.51 \\ 
        \textbf{Random} & 47.14 & 52.10 & 53.38 & 51.75 \\ 
        \bottomrule
    \end{tabular}
    \end{adjustbox}
\end{table}

The results in Table~\ref{tab1} validate the efficiency of the GODHS framework. Guided by both LLMs, our system demonstrates significantly lower search rates across all categories compared to the non-semantic baselines. This indicates that by leveraging hierarchical, semantic guidance, the robot needs to explore a much smaller fraction of the environment to locate the target, confirming that the approach significantly reduces search cost and improves efficiency.


We identified three categories of failure modes: (i) hardware limitations—for example, the manipulator cannot access areas close to the floor; (ii) insufficient common-sense in the LLM—for instance, it may instruct the robot to inspect a non-openable exterior panel of a fridge; and (iii) semantic ambiguity—such as misclassifying a billiard table.

\subsection{Performance Evaluation}\label{sec4.3}

In this section, we evaluate the performance of our heuristic-based motion planner, specifically the pose sorting strategies detailed in Sec.~\ref{sec3.3}. The experiment is designed to assess the effectiveness of our approach in optimizing the robot's exploration path and reducing execution time in the simulated environment.

To systematically assess the effectiveness of our proposed sorting strategies, we compare four configurations: an unoptimized baseline with no sorting applied to either EE or CH poses; a configuration where only EE poses are optimized via lexicographical sorting; another where only CH poses are optimized using polar angle sorting; and a final setup where both optimizations are applied together.

We evaluate three key performance metrics: the Normalized EE Path Length, which represents the ratio of the total EE travel distance to the theoretical shortest path; the Normalized CH Path Length, defined similarly for the CH poses; and the Execution Time Ratio, which is the ratio of execution time after optimization to that of the unoptimized case, with a lower value indicating better efficiency.

\begin{table}[htbp]
\centering
\caption{Comparison of Sorting Methods for EE and CH Poses.}
\label{tab3}
\begin{tabular}{lccc}
\toprule
\textbf{Sorting Method}   & \textbf{EE Ratio} & \textbf{CH Ratio} & \textbf{Time Ratio} \\
\midrule
Unoptimized & 2.81 & 2.37 & 1.00 \\
EE Sorting & \textbf{1.75} & 2.41 & 0.87 \\
CH Sorting & 2.79 & \textbf{1.60} & 0.83  \\
Both Optimized & \textbf{1.77} & \textbf{1.59} & \textbf{0.66}  \\
\bottomrule
\end{tabular}
\end{table}

Table~\ref{tab3} presents the comparative results. The data demonstrates that applying lexicographical sorting to EE poses and polar angle sorting to CH poses independently yield significant improvements in their respective path lengths. When both strategies are combined, the system achieves the highest optimization in both path efficiency and overall execution time, confirming the effectiveness of our heuristic motion planning strategy.

%% file: sections/sec5.tex
\section{Conclusion}

In this work, we presented the \textbf{GODHS Framework}, which integrates an LLM's commonsense reasoning with a multi-level decision process to improve search efficiency. This is achieved by using structured prompts to ensure reliable reasoning and a heuristic-based motion planner with \textbf{pose sorting} to generate efficient exploration trajectories. Experiments conducted in simulation demonstrated the feasibility of our approach and more efficient search performance compared to non-semantic strategies. Future work will focus on deploying the framework on a physical robot and exploring the integration of multi-modal foundation models~\cite{bommasani2022opportunitiesrisksfoundationmodels} and social navigation models~\cite{caicuriosity} to enhance its capabilities.